\documentclass[manuscript,nonacm]{acmart}
\AtBeginDocument{}

\setcopyright{acmlicensed}
\copyrightyear{2025}
\acmYear{2025}
\acmDOI{XXXXXXX}
\acmConference[HCXAI]{CHI Workshop on Human-Centered Explainability}{April 26--May 1,
  2025}{Yokohama, Japan}
\acmISBN{978-1-4503-XXXX-X/2018/06}

\usepackage{wrapfig}
\usepackage{soul}
\usepackage{parskip}
\begin{document}

\title{Thoughts without Thinking: Reconsidering the Explanatory Value of Chain-of-Thought Reasoning in LLMs through Agentic Pipelines}

\author{Ramesh Manuvinakurike}
\authornote{Authors contributed equally to this research.}
\email{ramesh.manuvinakurike@intel.com}
\author{Emanuel Moss}
%authornote{Both authors contributed equally to this research.}
\authornotemark[1]
\email{emanuel.moss@intel.com}
\author{Elizabeth Anne Watkins}
\authornotemark[1]
\email{elizabeth.watkins@intel.com}
\author{Saurav Sahay}
\email{saurav.sahay@intel.com}
\author{Giuseppe Raffa}
\email{giuseppe.raffa@intel.com}
\author{Lama Nachman}
\email{lama.nachman@intel.com}
\affiliation{%
  \institution{Intel Labs}
  \city{Hillsboro}
  \state{Oregon}
  \country{USA}
}

\renewcommand{\shortauthors}{Manuvinakurike et al.}

\begin{abstract}
  Agentic pipelines present novel challenges and opportunities for human-centered explainability. The HCXAI community is still grappling with how best to make the inner workings of LLMs transparent in actionable ways. Agentic pipelines consist of multiple LLMs working in cooperation with minimal human control. In this research paper, we present early findings from an agentic pipeline implementation of a perceptive task guidance system. Through quantitative and qualitative analysis, we analyze how Chain-of-Thought (CoT) reasoning, a common vehicle for explainability in LLMs, operates within agentic pipelines. We demonstrate that CoT reasoning alone does not lead to better outputs, nor does it offer explainability, as it tends to produce explanations without explainability, in that they do not improve the ability of end users to better understand systems or achieve their goals. 
  %We conclude with a proposal for how our agentic pipeline can be used in concert with CoT approaches to produce greater human-centered explainability for AI systems.
\end{abstract}

\begin{CCSXML}
<ccs2012>
 <concept>
  <concept_id>00000000.0000000.0000000</concept_id>
  <concept_desc>Do Not Use This Code, Generate the Correct Terms for Your Paper</concept_desc>
  <concept_significance>500</concept_significance>
 </concept>
 <concept>
  <concept_id>00000000.00000000.00000000</concept_id>
  <concept_desc>Do Not Use This Code, Generate the Correct Terms for Your Paper</concept_desc>
  <concept_significance>300</concept_significance>
 </concept>
 <concept>
  <concept_id>00000000.00000000.00000000</concept_id>
  <concept_desc>Do Not Use This Code, Generate the Correct Terms for Your Paper</concept_desc>
  <concept_significance>100</concept_significance>
 </concept>
 <concept>
  <concept_id>00000000.00000000.00000000</concept_id>
  <concept_desc>Do Not Use This Code, Generate the Correct Terms for Your Paper</concept_desc>
  <concept_significance>100</concept_significance>
 </concept>
</ccs2012>
\end{CCSXML}

%\ccsdesc[500]{Do Not Use This Code~Generate the Correct Terms for Your Paper}
%\ccsdesc[300]{Do Not Use This Code~Generate the Correct Terms for Your Paper}
%\ccsdesc{Do Not Use This Code~Generate the Correct Terms for Your Paper}
%\ccsdesc[100]{Do Not Use This Code~Generate the Correct Terms for Your Paper}

\keywords{Presented at the Human-centered Explainable AI Workshop (HCXAI) @ CHI 2025, DOI: \href{https://doi.org/10.5281/zenodo.15170393}{10.5281/zenodo.15170393}}

\received{28 February 2025}

\maketitle

\section{Introduction}

Explainable AI (XAI) has increasingly become acknowledged as a crucial, highly desirable property for developing and deploying AI systems responsibly~\cite{arrieta_explainable_2019}, accountably~\cite{smith-renner_no_2020}, and  effectively~\cite{ehsan_human-centered_2022}. Explanations are intended to help people understand and make appropriate use of AI systems' outputs, and to provide insight into systems' functionality to a range of interested parties~\cite{ehsan_human-centered_2022}. The need for such explanations is only heightened by the widespread interest in (and adoption of) AI systems based on large language models (LLMs) like the generative pretrained transformer (GPT) models central to ChatGPT~\cite{openai_introducing_2022} and DeepSeek~\cite{guo_deepseek-coder_2024}. Recently, model architectures have begun implementing ``chain-of-thought" (CoT) techniques~\cite{nye_show_2021, wei_chain--thought_2023} that prompt a model to "verbalize step-by-step reasoning"~\cite{turpin_language_nodate} and condition its output on that verbalization to improve outcomes~\cite{suzgun_challenging_2022, zelikman_star_nodate}. This CoT verbalization has been incorporated into frontier models as part of their `scratchpad' memory to do compositional thinking before generating an output~\cite{amiri2025lower}, and has also been proposed as a form of `CoT Explanations' or explainability, offering insight into the so-called reasoning process that lead to a particular output~\cite{wu2024usable} (but cf. ~\cite{li2024towards, turpin_language_nodate, tanneru2024difficulty}). 

While CoT can often resemble abductive reasoning~\cite{gabbay_integrating_2000, ignatiev_abduction-based_2019}, rationalizing and fitting likely explanations to given data, CoT text may often lead to \textit{explanations without explainability}, prone as it is to generating seemingly plausible but erroneous, contradictory, or irrelevant content~\cite{sanwal_layered_2025}. 
%\textcolor{violet}{In this paper, i) we introduce a novel agentic pipeline framework that can perform performance task guidance capable of performing between different types of requests by the users. ii) we demonstrate that CoT reasoning does not lead to better outputs, nor does it offer explainability for those outputs. 
%We present an empirical study of the former point and an analysis of the latter point, accompanied by a proposal for how our agentic pipeline can be used in concert with CoT approaches to produce greater explainability for AI systems.
%\textcolor{red}{@MANNY: I've edited this paragraph. Please let me know if it sounds ok:
In this paper, we introduce an agentic pipeline framework developed as part of a perceptive task guidance system to support factory technician task execution in a manufacturing setting
%~\footnote{This perceptive task guidance system uses multimodal AI approaches to observe tasks using multiple cameras, interact through speech and text with factory technicians, and guide technicians through a complex task-based protocol.}
, which is being tested and validated in a non-factory setting using the assembly and disassembly of toy vehicles (cranes, dump trucks, bulldozers, etc.) as a proxy for manufacturing tasks~\cite{sener2022assembly101}. The agentic pipeline framework we introduce below demonstrates that CoT reasoning does not lead to better outputs, nor does it offer explainability for those outputs. We present an empirical study of the former point and an analysis of the latter point, accompanied by a proposal for how our agentic pipeline can be used in concert with CoT approaches to produce greater explainability for AI systems.

\section{Related Work}

The emerging subfield of human-centered AI explainability builds on traditional explainability by broadening focus from AI models to include sociotechnical components of AI deployment. This includes consideration of how non-expert AI end-users understand how AI systems work, how inputs are created, and how explanations inform user behaviors \cite{kim2023help}. The emergence of LLMs has complicated this field, as explored in prior iterations of the CHI HCXAI workshop \cite{ferrario2024addressingsocialmisattributionslarge,zytek2024llmsxaifuturedirections,balayn2024understandingstakeholdersperceptionsneeds,sarkar2024largelanguagemodelsexplain}. %While LLMs can provide written text that appears to "explain" why and how it produced its outputs, this content, like most other LLM outputs\cite{kim2024m}, are subject to inaccuracy and overconfidence\cite{ye2022unreliabilityexplanationsfewshotprompting}. %This emergence combines all the complexity of human-centered XAI %(with its recognition that sociotechnical parameters shape how a technology is perceived, understood, and used),
%with all the challenges of how best to provide to people with explanation that cannot necessarily explain. 
%Exacerbating these challenges is the ``black box" character of LLMs, which obscure how well designers and end users can understand the relationship of inputs to outputs, the chains of prompts that agents produce to pursue tasks, and any explanations rendered by AI systems about their behaviors~\cite{burrell_how_2016}.  
While we are still in the early days of untangling this knot, agentic systems have been proposed as a means to produce explainability~\cite{nooralahzadeh_explainable_2024, sanwal_layered_2025, wang_survey_2024,ehsan2019automated}, in that they can produce a record of how instructions are passed between agents, ``allowing end users to trace conclusions back to their source data"~\cite{nooralahzadeh_explainable_2024}. (See Section 4 for more detail on this architecture.) However, agentic architectures risk producing, instead of explainability, crowds of LLMs playing a sort of children's game of "telephone" in which people whisper what they think is the same message from person to person in a chain until the last person says the message and everyone finds out it has been inevitably distorted, in often funny ways, by its passage between multiple people. These crowds of LLM agents present new challenges of "cascading risks," \cite{mitchell2025fully} where humans have ceded control and oversight to information passing through a multi-actor system\cite{chan2023harms}.

%Manny: Relevant stuff explainability \& agentic work

\section{Data}

We prepared a benchmarking dataset to evaluate the answering performance of various models within the agentic pipeline. This dataset consists of two categories of questions: task-based (Task) questions and organizational/social (Org-Soc) questions. We modeled our task-based questions on the Assembly 101 dataset \cite{sener2022assembly101}, which provided an open-source way to build a system that supports human performance of physically situated tasks. 
Our Task questions revolve around putting together and taking apart toys like cranes, trucks, excavators, etc. and were used to probe the model about task-based performance support (e.g., How do I remove the wheels?). For our Org-Soc questions, we took a novel participatory approach\cite{manuvinakurike2024qa}. We interviewed manufacturing technicians with whom we collaborate building our perceptive task guidance system. We worked with them to compile a set of questions they would want to ask the agent in the course of their real-world work. This process revealed that technicians assume that AI assistants like our system has a far wider range of abilities than they actually do. This insight (forthcoming) points to a gap between the way that foundation models' performance is measured using existing benchmarks, and the real world of complex overlap between tasks, organizations, and social reality into which LLMs and agentic systems are being deployed. This participatory approach brings the evaluation of our system 
%LLMs (and agentic pipelines constructed with LLMs) 
closer to the real world.
Our dataset from toy assembly consists of 152 participatory Org-Soc questions, 43 Task questions per toy (3 toys evaluated).
%, building a more robust testing procedure than other benchmarking datasets that typically rely on speculative questions. 
Both sets of questions (Task and Org-Soc) were posed to multiple models. The responses (N=750/300 (Task), 300 (Org-Soc), 150 (Thoughts)) were then scored along a Likert-like scale (-1, 0, 0.5, 1)\footnote{On this scale, ``1" was a correct answer, ``0.5" was a partially correct answer or an answer that had correct content in it but also included unhelpful or confusing information, ``0" was an incorrect answer, and ``-1" was a dangerous or unsafe answer.} for their accuracy, comprehensiveness, and helpfulness, using an idealized technician performing a required task as the entity for whom an answer should assist. The scoring was first conducted on a subset of question-CoT-answer tuples by three authors, who all scored the same sub-subset of tuples to coordinate inter-annotator agreement, and then LLM-as-a-judge (GPT-4o\cite{openai2024gpt4o}) was prompted with the scoring scheme, which scored the tuples (N=2529). 
%The remaining responses were annotated (annotators were unaware of the model that generated the response) by the authors. 
Alongside this numerical scoring and quantitative analysis, we also conducted a qualitative analysis of the CoT texts in relation to questions and answers. %  (Section \ref{sec:Discussion}). 

%\textcolor{red}{RAMESH: discussion of more data processing, if needed, could go here. Could also mention the JSON that allowed us to inspect the entire pipeline, but that is somewhat outside the scope of the paper.}

\section{Agentic architecture}

%The system we implemented, which is capable of performing task guidance for human activities, is agentic. 
The allure of agentic architecture is its ability to decompose the task (using planners), leveraging external systems (tools  \& other agents),  perception, memory components and different task action experts (e.g, RAG, Reformulator etc.). Our implementation consists of: i) Perceptors: Perceive visual and language data, ii) Planners: tasked with decomposing, iii) Actors: Expand context, generate response and verify safety. 
%All the agents (with the exception of Visual action recognition, object detection) are LLM based.  
%(Table~\ref{tab:agentic_details} contains full list). 
Upon receiving the inputs from the perceptors, the system's lead planner sequences the agents \& tools invocations. Depending on the input (e.g., Org-Soc or Task), the planner decides to invoke the RAG (Retrieval Augmented Generation)-based flow or Chit-chat module. A detailed description of the system implementation can be found in Appendix~\ref{sec:appendix_agent_implementation}. 
The answer planner is then tasked with planning if there is sufficient information to answer the question, or if a follow-up question needs to be generated. The safety agent performs the final check.
%and the response is generated if safe. Else, the response is regenerated by the response generation module until . -- Beyond scope of this work. 

%\subsection{LLMs}
Each agent in the agentic flow contains its own LLM. We experiment with 3 different LLMs (Llama3 - 8b \cite{dubey2024llama}, Qwen - 7b, 14b \cite{qwen}) and their reasoning based Deepseek distilled alternatives \cite{deepseekai2025deepseekr1incentivizingreasoningcapability}. Deepseek distilled reasoning models achieve impressive performance in numerous tasks while generating CoT reasons \cite{deepseekai2025deepseekr1incentivizingreasoningcapability}. In this work, we aim to analyze the application of the models for our task guidance application and the unique dataset we developed, which requires better context understanding and complex reasoning. The context for the answer generation consists of lengthy specification documents (instructional documents detailing the steps needed carry out the process), instructions, historical chat data, and additional data from supporting agents. 
The questions from the users often requires reasoning and/or searching for a needle in a haystack \cite{kuratov2024babilong,anthropic2023long}. 
The goal of our work is to evaluate the responses generated by the agentic system.
%with the LLMs mentioned on their relevance to the task or Org-Soc constraints. 

%The reasoning thoughts generated by the models helps achieves explainability goals while helping models generate better answers. 
%Our work aims to study these models in a task guidance scenario.% on participatory dataset. 
%In this work, we focus on using models without finetuning. 

\begin{table}[]
    \centering
%    \resizebox{\textwidth}{!}{
    \begin{tabular}{p{0.05\linewidth} | p{0.95\linewidth}}
        Tuple & Content \\ 
        \hline
         & \scriptsize{\textbf{Input:} What is the dump box connected to?} \\ 
        A  & \scriptsize{\textbf{CoT:} Okay, so I'm trying to figure out what the "dump box" is connected to in this context. From the conversation history, there's a mention of a guide detailing a 152-step process for disassembling and reassembling a toy. The user specifically asks about the dump box connected to something.  First, I need to understand what a ``dump box" typically refers to. In mechanical terms, a dump box is often part of a motor or transmission system. It's usually a component that holds clutch plates or other parts that need to be engaged or disengaged rapidly. The dump box is connected to the clutch system and operates with solenoids to engage or release the clutch plates, which allows the vehicle to shift gears.  Looking back at the guide mentioned earlier, it involves taking apart various components of a mechanical toy, possibly similar to a robot or an RC car. The steps include disassembling parts like the crane arm, transmission, and other mechanical systems.  Since the user is asking about the dump box specifically, I should focus on its connection points. In a typical setup, the dump box is connected to the motor via a transmission system. It might also be linked to the gear shifter or control mechanism of the toy.  In the context of the guide, during disassembly, one would need to disconnect the dump box from the clutch and other components before taking them apart further. During reassembly, the process would reverse, ensuring the dump box is correctly reattached and aligned with the rest of the system for proper functionality.  So, putting this together, the dump box in this context is likely connected to the motor and transmission systems of the toy. It's a crucial component for shifting gears and controlling the movement of the toy's mechanical parts.} \\ 
           & \scriptsize{\textbf{Output:}   In the context of the detailed guide for disassembling and reassembling a complex mechanical toy, the ``dump box" is a component typically found in the transmission system. It is connected to both the motor and the clutch system, allowing for rapid engagement or disengagement of the clutch plates. This connection ensures proper gear shifting and control over the movement of the toy's mechanical parts during operation.  **Summary:** - **Dump Box Function**: Part of the transmission system, responsible for engaging/disengaging clutch plates. - **Connection Points**: Linked to the motor and other subsystems like the gear shifter. - **Disassembly/Reassembly**: Requires careful handling to ensure correct reattachment and functionality during reassembly.} \\ \hline
         & \scriptsize{\textbf{Input:} How do you ensure data privacy and security?} \\ 
         B & \scriptsize{\textbf{CoT:} Okay, so I'm trying to figure out how an AI like MARIE ensures data privacy and security. I know that as a helpful agent developed at Intel Labs, it's probably designed with some safeguards in place. But I'm not exactly sure about the specifics.  First, maybe there are layers of encryption. That makes sense because when you send information, especially sensitive stuff, you want it protected during transit. So if MARIE uses HTTPS or something similar, that could be part of it.  Then, access controls come to mind. Only authorized people should handle data, so maybe there are strict permissions and role-based access. That way, even if someone gets in, they can't access everything without proper clearance.  Data anonymization is another point. Maybe personal information is removed or obscured before processing. That way, the data isn't identifiable even if it's intercepted.  Regular audits might also be part of the process to check security measures and ensure compliance with regulations like GDPR or CCPA. This helps keep things up to standard.  Moreover, secure development practices must be in place. Following best practices when coding ensures that any vulnerabilities are caught early, making the system more secure overall.  Physical security isn't something I initially thought about, but if MARIE is deployed in a physical space, securing servers and data centers from unauthorized access would be crucial.  Lastly, continuous monitoring and incident response plans can catch any breaches quickly. This proactive approach helps minimize damage and recover faster if something goes wrong.  I wonder how these measures are implemented technically, like specific algorithms or protocols they use for encryption. Also, how they handle updates and patches to maintain security over time.} \\ 
          & \scriptsize{\textbf{Output:}   MARIE ensures data privacy and security through a multi-layered approach:  1. **Encryption**: Utilizes HTTPS and other secure communication protocols to protect data during transit.  2. **Access Controls**: Implements strict permissions and role-based access to restrict data handling to authorized personnel only.  3. **Data Anonymization**: Removes or obscures personal information before processing, making data non-identifiable even if intercepted.  4. **Regular Audits**: Conducts compliance checks with regulations like GDPR and CCPA to maintain standards and ensure proper practices.  5. **Secure Development Practices**: Follows best practices in coding to identify and mitigate vulnerabilities early, enhancing overall security.  6. **Physical Security**: Secures servers and data centers against unauthorized access, ensuring physical protection.  7. **Continuous Monitoring and Incident Response**: Monitors for breaches and has plans in place to respond quickly, minimizing potential damage and recovery time.  These measures are technically implemented through algorithms and protocols, with ongoing updates and patches to maintain security over time.} \\ \hline 
           & \scriptsize{\textbf{Input:} Can you communicate with humans in multiple languages?} \\
        C & \scriptsize{\textbf{CoT:} Okay, so I'm trying to figure out how an AI like MARIE ensures data privacy and security. I know that as a helpful agent developed at Intel Labs, it's probably designed with some safeguards in place. But I'm not exactly sure about the specifics.  First, maybe there are layers of encryption. That makes sense because when you send information, especially sensitive stuff, you want it protected during transit. So if MARIE uses HTTPS or something similar, that could be part of it.  Then, access controls come to mind. Only authorized people should handle data, so maybe there are strict permissions and role-based access. That way, even if someone gets in, they can't access everything without proper clearance.  Data anonymization is another point. Maybe personal information is removed or obscured before processing. That way, the data isn't identifiable even if it's intercepted.  Regular audits might also be part of the process to check security measures and ensure compliance with regulations like GDPR or CCPA. This helps keep things up to standard.  Moreover, secure development practices must be in place. Following best practices when coding ensures that any vulnerabilities are caught early, making the system more secure overall.  Physical security isn't something I initially thought about, but if MARIE is deployed in a physical space, securing servers and data centers from unauthorized access would be crucial.  Lastly, continuous monitoring and incident response plans can catch any breaches quickly. This proactive approach helps minimize damage and recover faster if something goes wrong.  I wonder how these measures are implemented technically, like specific algorithms or protocols they use for encryption. Also, how they handle updates and patches to maintain security over time.} \\
         & \scriptsize{\textbf{Output:} Currently, [the agentic pipeline system] is only capable of communicating in English.} \\ \hline
    \end{tabular}
%    }
    \caption{Example tuples. Input is from the Task and Org-Soc participatory benchmark dataset, chain-of-thought and output are produced by the agentic pipeline.}
    \label{tab:tuples}
\end{table}

\begin{figure}
    \centering
    \includegraphics[width=0.8\linewidth, trim=0 13.5cm 0 0, clip]{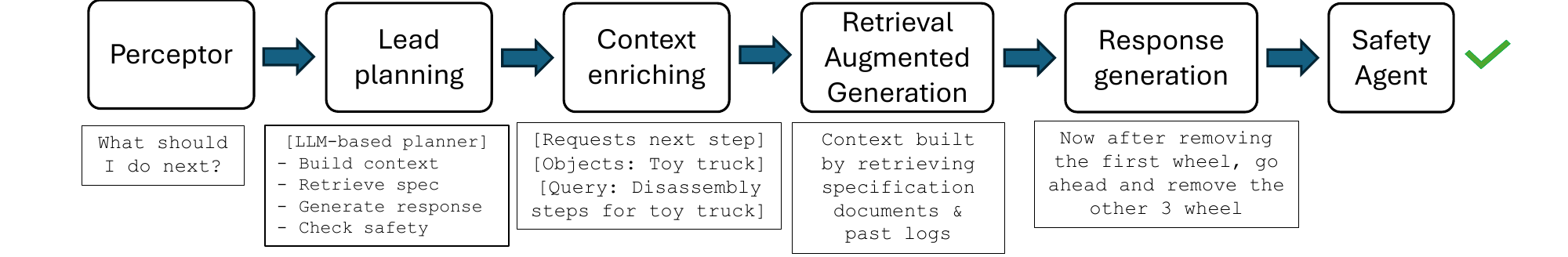}
    \caption{Shows sample agentic flow depending on the type of input query from the user. Once the question is ingested, the lead planner generates a plan which is a sequence of agentic calls. The output of each of the agent in the flow is shown here.
    %The response generation is further a Router agent which either calls a question generator (if information available is insufficient to answer the question) or the answer generator.  
    }
    \label{fig:agentic_flow}
\end{figure}

\section{Experiments}

We conduct the experiments to evaluate the agentic architecture offline. 
The questions are input to the systems where agentic flow generate the responses. The responses are then reviewed and scores assigned by expert human reviewers \& LLM-as-a-Judge. Since ground truth answers are rather difficult to generate and compare to the system generated answers, we adopt human \& reference-free LLM-Judge evaluation. The experts assign scores to every answers generated by the agentic flow using 6 different LLMs. 
%We also rate the responses using LLM-as-a-judge utilizing the best practices while acknowledging the limitations of the approach \cite{gu2024survey}. 
The reviewer scores are thus assigned by human experts and LLMs-as-a-judge. 
To further analyze the thoughts generated, we generate the scores for the thoughts (both human reviewers and LLM-as-a-judge) . 
The reviewers are tasked to assign 4 scores to the answers using question, reference response (and spec documents for task based questions) as context (see Footnote 1 for the judgment scale). %1 for the correct answer, 0.5 for partly correct answer (if the response contains some information not relevant or incorrect), 0 for incorrect answer and -1 if the response is dangerous or unsafe. 

\section{Results} \label{sec:Results}

\begin{figure}
    \centering
    \includegraphics[width=0.9\linewidth, trim=0 8cm 0 0, clip]{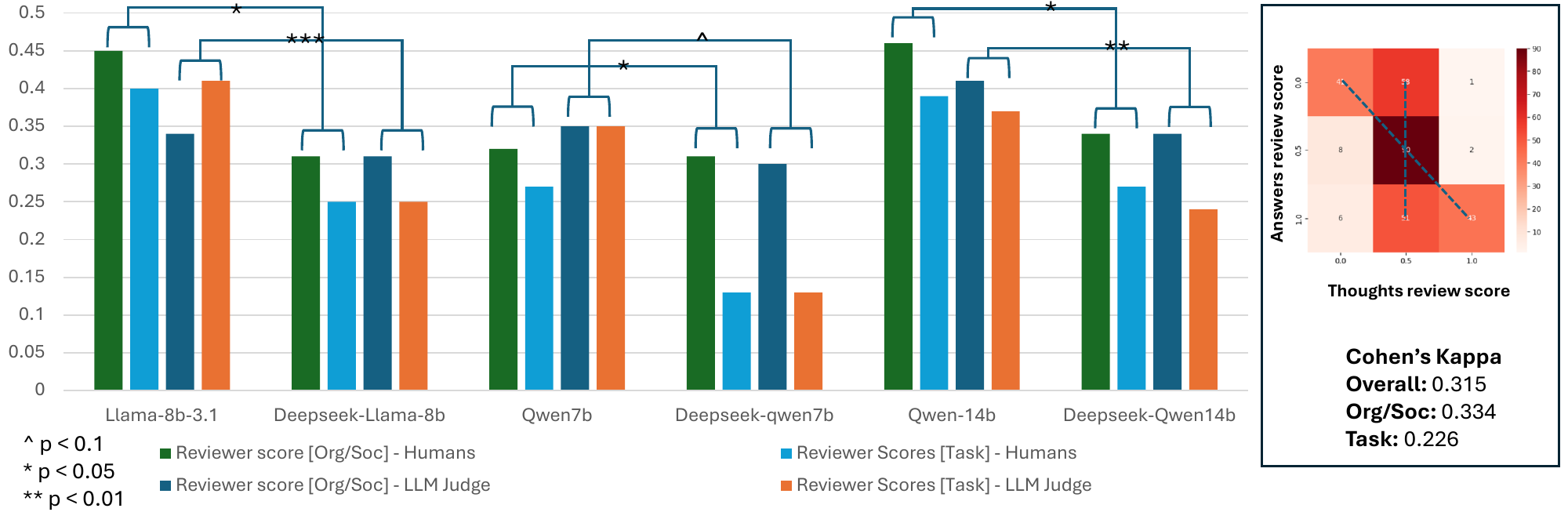}
    \caption{Shows the reviewer scores for the answers as rated by humans and LLM-as-a-judge on Task and Org-Soc questions. We observe that the reviewer scores for the non-reasoning models are better than their reasoning (Deepseek-) counterparts. We can also observe that the thoughts reviewer scores are weakly correlated with the answer reviewer scores. }
    \label{fig:results}
\end{figure}

%Figure~\ref{fig:results} shows the results. 
\textbf{Non-reasoning models perform better}: We find that the thought models' generated response (Deepseek distilled models-7b, 8b, 14b) achieve lower reviewer scores than the models not generating/trained to generate thoughts. We perform Wilcoxon rank-sum test and show that the differences are significant (Figure~\ref{fig:results}). \textbf{LLM-as-a-Judge or not}: Usage if LLM-as-a-Judge remains an area of debate \cite{zheng2023judging,tan2024judgebench,gu2024survey}. In this work, we find that the LLM Judge \& human reviewer scores exhibit weak correlation (Cohen's Kappa = 0.48 (org-soc), 0.28 (task)). We find that the LLM-as-a-Judge, in spite of its limitation, can be a reference-free approach that could serve as an indicator of the quality of the response. \textbf{Reviewer scores for the answers in reasoning models are weakly correlated with the thoughts} Interestingly, we find that the reviewer scores for the thoughts are correlated with reviewer scores for the responses. This indicates that the thoughts do not necessarily guide the models to the correct responses. If the thoughts were strongly correlated to the answers, we might infer that the "correctness" or "incorrectness" of the answers could explained by the thoughts. However, we observe that the thoughts have high degree of incompleteness (0.5 reviewer score) irrespective of the answers being correct or wrong (middle column in the heatmap Figure~\ref{fig:results}). We highlight this as an interesting area for future exploration. 

\section{Discussion} \label{sec:Discussion}
The data presented above indicates that CoT can lead to incorrect answers, hampering explainability. To investigate these implications, we also conducted a qualitative content analysis (QCA) ~\cite{kracauer_challenge_2025, moretti_conjectures_2000} of CoT reasoning. QCA can act as a form "reverse engineering"~\cite{smith_what_2016} that allows analysts to focus on the entire system that produced outputs. %or much smaller (e.g., specific phrases, word choice, metaphors, and analogic relationships)~\cite{moretti_conjectures_2000}.
Take, for example, the prompt-CoT-output tuple A in Appendix Table \ref{tab:tuples}. Given the input, the output is unhelpful and incorrect. It contains references components that are not part of a toy dump truck (e.g., transmission system, clutch) that are not mentioned in the assembly instructions. From an inspection of the CoT, it does not offer users an explanation for why the output is so unhelpful, but \emph{it does show how quickly the CoT reasoning went astray}. 

A QCA of the CoT, however, does provide some clues as to what may have led it astray by pointing to larger elements than the tuple itself. A qualitative reading of the CoT shows how quickly the text moves away from references to ``disassembling and reassembling a toy"  (as provided by the assembly instructions) and toward references to what is more ``typically" associated with a dump box: an actual machine. Here, the CoT pulls in a range of tokens related to machine components like a ``clutch", ``transmission", or ``gears" but while it continues to mention the ``toy", nouns and verbs related to actual machines predominate. Put together, one hypothesis generated by QCA points toward the tendency of LLMs to fall victim to the ``Einstellung Paradigm"~\cite{naeini_red_herring} in which a focus on familiar or common approaches leads away from the correct approach. Here, the CoT focuses on tokens much more commonly associated with dump trucks (machine-related) in the LLM rather than the tokens germane to the task (toy-related and provided in the RAG). Another possible hypothesis generated by a QCA is that the RAG agent seemed to face an issue in filling up the context window, and therefore drew on the foundation model, which returned text on dump trucks, generically, as complex machinery. Crucially, while the QCA has pointed us in potentially helpful directions for troubleshooting this unhelpful output, such detailed analysis of the CoT does not lead to explainability. Rather, considerably more effort is required to make sense of inputs and outputs, even if the CoT generates more material with which to engage in such sensemaking. Tuple B in Appendix Table \ref{tab:tuples} also evinces a susceptiblity to the Einstellung paradigm.

Another way explainability is hindered by CoT is seen in tuple C in Appendix Table \ref{tab:tuples}. The input is a question from the Org-Soc questions, ``Can you communicate with humans in multiple languages?" that indexes a need users might have to know how best to communicate with the system. The output, stating that it cannot, is incorrect, even though a test prompt in a non-English language returns a legitimate result in that other language.\footnote{The input "Wo wohnen sie?" ("Where do you live?") returns an output stating that, as an AI, our task-based guidance system cannot be thought of as having a traditional place of residence. Interestingly, the CoT for this prompt is output in English.} Here, the CoT text illustrates three important ways that CoT can materially hinder explainability: 1) It takes as a fact something that is not true even though no prompt provided as context to an input states this, although it is true that all prompting is \textit{in} English. 2) It makes a flawed inference based on this claim: ``Since the context only lists English... right now it's just English", evincing a form of logical fallacy known as \textit{hasty generalization}~\cite{walton_rethinking_1999}. 3)The CoT produces more text than is needed to arrive at its (erroneous) conclusion, introducing difficult-to-parse text that the user needs to wade through in search of a useful explanation and also reinforces logical errors. Each of these interferes with explainability by introducing irrelevant text or erroneous forms of argument, raising the burden for anyone seeking explanatory content in the CoT.

\textbf{Limitations and Future Work: }
There are several limitations of our work. We assume agentic architecture as a preferred choice for developing the system while there exists a plethora of ways to developing the system. We aim to compare and evaluate alternate ways to develop task guidance system and evaluate via ablations the value of each agent in the system. We deploy the same LLM in each of our agent for ease of comparison. We acknowledge that different LLMs could serve as a different task experts. Our dataset while participatory in nature is limited in size. The aim of this work was to carefully evaluate every response from every model. In the future work, we aim to address this issue by scaling the analysis to a larger sample. 
%\textbf{Why thoughts lead to the incorrect answers}

%\section{Future Work}

\bibliographystyle{ACM-Reference-Format}
\bibliography{sample-base}

\appendix

\begin{figure}
    \centering
    \includegraphics[width= \linewidth]{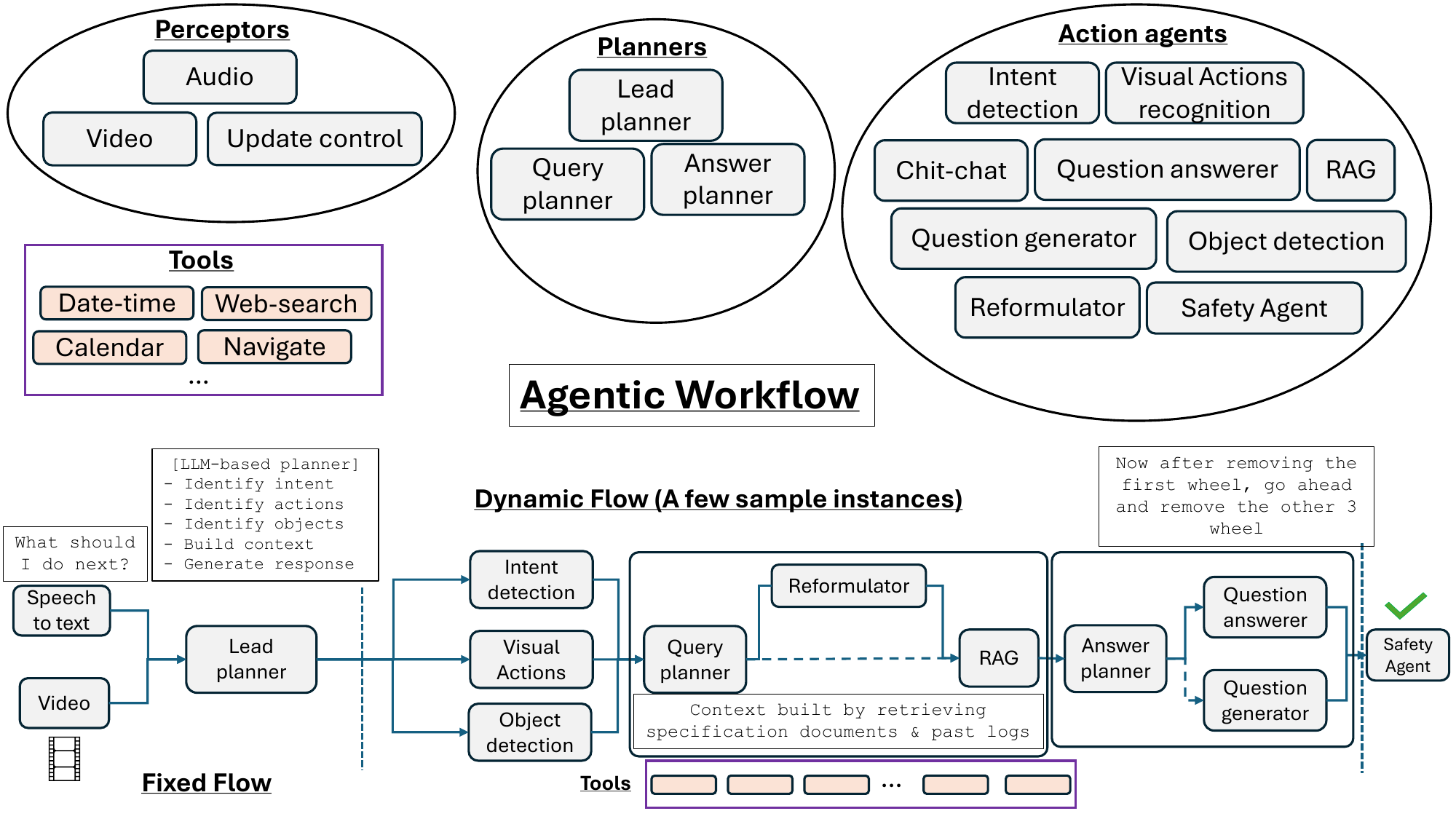}
    \caption{The figure shows our agentic implementation of the perceptive task guidance system. The agents are categorized into perceptors, planners and action agents. The agents are autonomous and rely either on LLMs or alternative deep learning models to accomplish their task. The agentic pipeline consists of fixed flow and dynamic flows. The dynamic flows involves planner invoked components. The plan generated by the planner is converted into agent invocations (Routers). }
    \label{fig:agentic_flow}
\end{figure}

\section{Agentic implementation details}
\label{sec:appendix_agent_implementation}

\begin{figure}
    \centering
    \includegraphics[width=\linewidth]{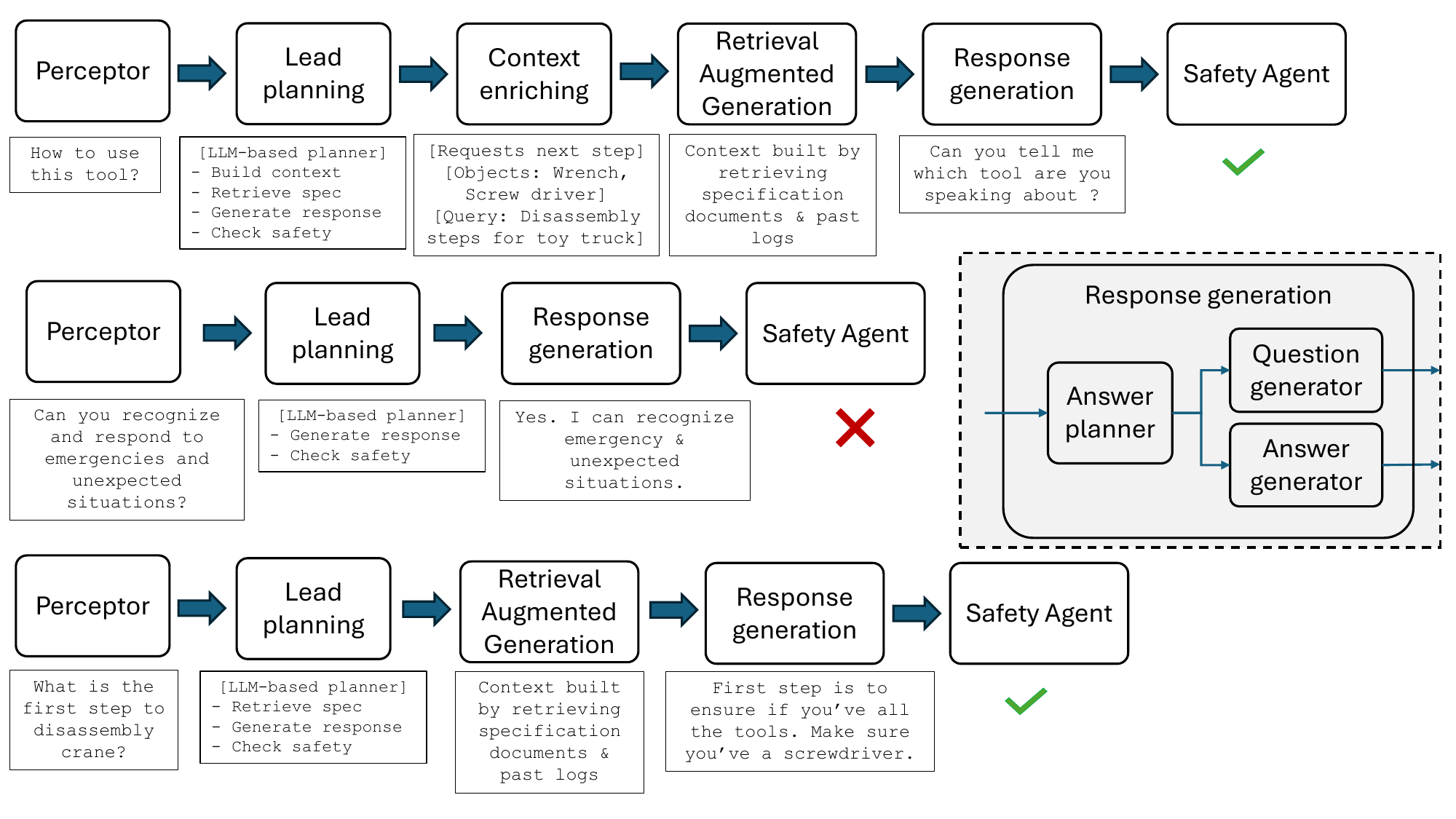}
    \caption{Shows more examples of dynamic flows in the agentic flow. The examples show the input question from the users in the Perceptor. The Lead planner creates the plan and creates the Route which consists of agent calls. RAG consists of spec documents for each toy which is further chunked. The RAG module further converts the input context into a query to the database. The retrieved document along with the inputs is passed to the following agents. The Response generation module consists of planners which decides if the context is sufficient to answer the question or not. The answer planner then invokes the expert question generator or the answer generator. The responses are always verified before publishing the responses to the users. }
    \label{fig:enter-label}
\end{figure}

\begin{table}
    \centering
    \begin{tabular}{l|p{6cm}|p{4cm}}
        \textbf{Agent} & \textbf{Function} & \textbf{Input} \\ \hline
         Lead planner & Fixed lead component tasked with creating a pipeline graphs which each node is an agent. & Definitions of the agents, input (question), and task description \\ \hline
         Query planner & Assesses the query and decides if the query needs to be reformulated and which database to use for retrieval augmented generation & Task description, Original query, Additional text context (output of trailing agents) \\ \hline
         Answer planner & Decides if the context has sufficient information to answer the question. If it does, the answer planner is invoked to generate the answer else, a question generator. & Task description, context (output of trailing agents), query \\ \hline
         Intent detection & Detects the intent (defined by domain experts) of asking the question. This can be especially useful for either the RAG module or other subsequent module. & Task description, intent candidates, Original query  \\ \hline
         Visual action recognition & Identifies the action being performed in the visual frame & Vision frames \\ \hline
         Object detection & Identifies objects in the visual frame & Vision frames \\ \hline
         Chit-chat & For non-task related queries, the system answers the queries following a set of policy rules defined in the guideline prompt & Task description, Policy document, Query\\ \hline
         Question answerer & Answers the task-related query utilizing the enhanced context & Task description, Query, context \\ \hline
         Question generator & Generates the question typically requesting for missing information utilizing the query and the context & Task description, Query, context \\ \hline
         Reformulator & Converts the original query into a paraphrase query utilzing additional context parameters (typically objects and the actions) &  Task description, Query, Visual context \\ \hline
         RAG & Retrieves the documents using the query and summarizes the documents for the context utilizing the query & Task description, Query, Vector database\\ \hline
         Safety Agent & Utilizes a policy document to distinguish safe from inappropriate response & Policy document, response (question or answer) \\ \hline
    \end{tabular}
    \caption{Table describes the agents, functions and inputs to each agent.}
    \label{tab:agentic_details}
\end{table}

\begin{table}[]
    \centering
    \resizebox{\textwidth}{!}{
    \begin{tabular}{l|l}
            \textbf{Task examples} & \textbf{Org-Soc examples} \\ \hline
            What is the bull bar?&  What is your name and model number?\\
            How many pieces are there, in total?&  Can you tell me more about your capabilities and limitations? \\ 
            Which way do I turn the screws to unscrew? & Can you communicate with humans in multiple languages?\\
            Where is the windshield?& How fast can you process information? \\
            Which side of the dump box faces out?&  How do you ensure your own maintenance and self-preservation?\\
            What position should the grill bar be in?& Can you understand and process human language as easily as a human can? \\
            How tight should the nuts be?& How do you ensure privacy and security of my data? \\
            What is the correct way to remove the nuts?&  How do you ensure that you do not misunderstand a my request or need?\\ 
            \hline
            
    \end{tabular}
    }
    \caption{A few examples from our dataset that the system was evaluated on. }
    \label{tab:examples}
\end{table}

% %\textcolor{green}{TO BE DELETED: just throwing this image in - somewhere in the paper if we can add the benchmarking process visually .tasks, reviewing by humans vs. judges, is there any reference?, can we automate the pipeline to generate Poor quality explanations to Good quality ... ? .. something like this:}
% \begin{figure}
%     \centering
%     \includegraphics[width=0.5\linewidth]{image.png}
%     \caption{TO BE DELETED: Benchmarking Tasks and Evaluation}
%     \label{fig:enter-label}
% \end{figure}

\end{document}